\documentclass{article}
\usepackage{spconf_new,amsmath,graphicx}
\usepackage{subcaption}
\usepackage{multirow}
\usepackage{xcolor}
\usepackage{hyperref}

\usepackage{tablefootnote}

\newcommand{\rulesep}{\unskip\ \vrule\ }
\title{Guided contrastive self-supervised pre-training \\for automatic speech recognition}
\name{
\begin{tabular}{@{}c@{}}
Aparna Khare$^{1}$,
Minhua Wu$^{1}$,
Saurabhchand Bhati$^{2\star}$,
Jasha Droppo$^{1}$,
Roland Maas$^{1}$
\end{tabular}}
\address{$^{1}$Amazon Alexa, USA \\$^{2}$Johns Hopkins University, USA
 \vspace{-2mm}
\thanks{*The author contributed to this work during an internship at Amazon}}
\begin{document}
%
\maketitle
\begin{abstract}
Contrastive Predictive Coding (CPC) is a representation learning method that maximizes the mutual information between intermediate latent representations and the output of a given model. It can be used to effectively initialize the encoder of an Automatic Speech Recognition (ASR) model. We present a novel modification of CPC called Guided Contrastive Predictive Coding (GCPC). Our proposed method maximizes the mutual information between representations from a prior-knowledge model and the output of the model being pre-trained, allowing prior knowledge injection during pre-training. We validate our method on 3 ASR tasks: German, French and English. Our method outperforms CPC pre-training on all three datasets, reducing the Word Error Rate (WER) by 4.44\%, 6.55\% and 15.43\% relative on the German, French and English (Librispeech) tasks respectively, compared to training from scratch, while CPC pre-training only brings 2.96\%, 1.01\% and 14.39\% relative WER reduction respectively.

\end{abstract}
\section{Introduction}
\label{sec:intro}
Self-supervised Learning (SSL) has drawn a lot of recent attention in the machine learning community. After its successful applications in the natural language processing domain \cite{devlin2019bert, peters2018deep, radford2018improving}, it has also become an active research area for speech processing. 

One of the main categories of SSL methods learns representations by reconstructing the signal such as full reconstruction with autoencoders \cite{chen2019audio, chorowski2019unsupervised}, future reconstruction with Autoregressive Predictive Coding (APC) \cite{chung2019unsupervised} and masked reconstructions \cite{liu2020mockingjay, ling2020deep, ling2020bertphone}. Instead of reconstructing the exact signal, HuBERT \cite{hsu2021hubert} learns representations by utilizing an offline clustering step to provide aligned target labels for a masked prediction loss. Another category of SSL technology in literature learns representations through a contrastive loss by distinguishing a true future audio sample from a set of negative examples, such as the Contrastive Predictive Coding (CPC) model \cite{oord2018representation} and wav2vec \cite{schneider2019wav2vec}. Vq-wav2vec \cite{baevski2019vq} uses a vector quantization module in addition to contrastive loss to learn discrete representations and wav2vec 2.0 \cite{baevski2020wav2vec} minimizes the contrastive loss defined over contextual representations in the masked region. In addition, w2v-BERT \cite{chung2021w2v} combines the two categories by optimizing two self-supervised losses simultaneously (the contrastive loss and masked language modeling loss).

All of these methods learn representations from the acoustic data distribution only, which may not be optimal for the downstream ASR task. More recently, Wang et al. propose two supervision-guided codebook generation approaches to get better pre-trained embeddings for the downstream ASR task in \cite{wang2022supervision}. On top of HuBERT pre-training, it uses the phoneme alignments as training targets. It also tries to perform K-means clustering on the supervised speech features extracted from an end-to-end CTC model \cite{graves2006connectionist}. However, this work focuses on the masked prediction self-supervised learning and all the ASR experiments are conducted with the Librispeech dataset with just a few hundred hours of labeled data. In our work, we focus on exploring the contrastive loss based SSL method instead and experiment with large-scale datasets. We propose to introduce weak guidance to improve alignment between the learned representations and the downstream task. The weak guidance is provided in the form of posteriors from a prior-knowledge model learned from a small labeled dataset, which will be discussed in detail in Section \ref{sec:method:gcpc}.

To combine the self-supervised and supervised training to improve performance of the final ASR task, most existing methods in the  literature adopt a 2-stage scheme, where only the self-supervised loss is optimized at the first pre-training stage, and the supervised loss is optimized at the second stage. Wav2vec \cite{schneider2019wav2vec} and vq-wav2vec \cite{baevski2019vq} build the wav2letter \cite{collobert2016wav2letter} acoustic model by using the pre-trained embeddings as input features instead of log-mel filterbanks. Wav2vec 2.0 \cite{baevski2020wav2vec} and HuBERT \cite{hsu2021hubert} pre-train the transformer based encoder using the self-supervised loss, add a randomly initialized output layer on top and fine-tune with the CTC loss \cite{graves2006connectionist}. More recent research has shown that joint training with both supervised and unsupervised losses during the pre-training/fine-tuning stage or as a single training process helps improve the ASR performance. The initial UniSpeech work \cite{wang2021unispeech} demonstrates that representations learned during pre-training can be improved if the self supervised contrastive loss is combined with phonetic CTC loss, and the following Unispeech at scale work \cite{wang2021unispeech2} demonstrates better representations from the pre-training stage for the downstream ASR task when combining the contrastive loss and the transducer loss. \cite{talnikar2021joint} alternatively minimizes an unsupervised masked CPC loss and a supervised CTC loss. This single-stage method is shown to
match the performance of the two-stage wav2vec 2.0 on the Librispeech 100-hours dataset. \cite{raghavan2021hybrid} uses multitask learning comprising of supervised CTC, attention
and self-supervised reconstruction losses to directly train acoustic models under low-resource settings. \cite{bai2022joint} explores the benefit of combining the supervised RNN-T loss \cite{graves2012sequence}, the self-supervised contrastive loss and masked language modeling (MLM) losses during different training stages. In this paper, we demonstrate benefits of our proposed method mainly on the conventional 2-stage training scheme. We additionally try the joint training scheme on one ASR task during the ablation study and demonstrate gains similar to what is reported in literature.


\section{Method}
\label{sec:method}
\subsection{Contrastive predictive coding}
\label{sec:method:cpc}
\begin{figure}[h!]
\centering
  \includegraphics[width=\linewidth]{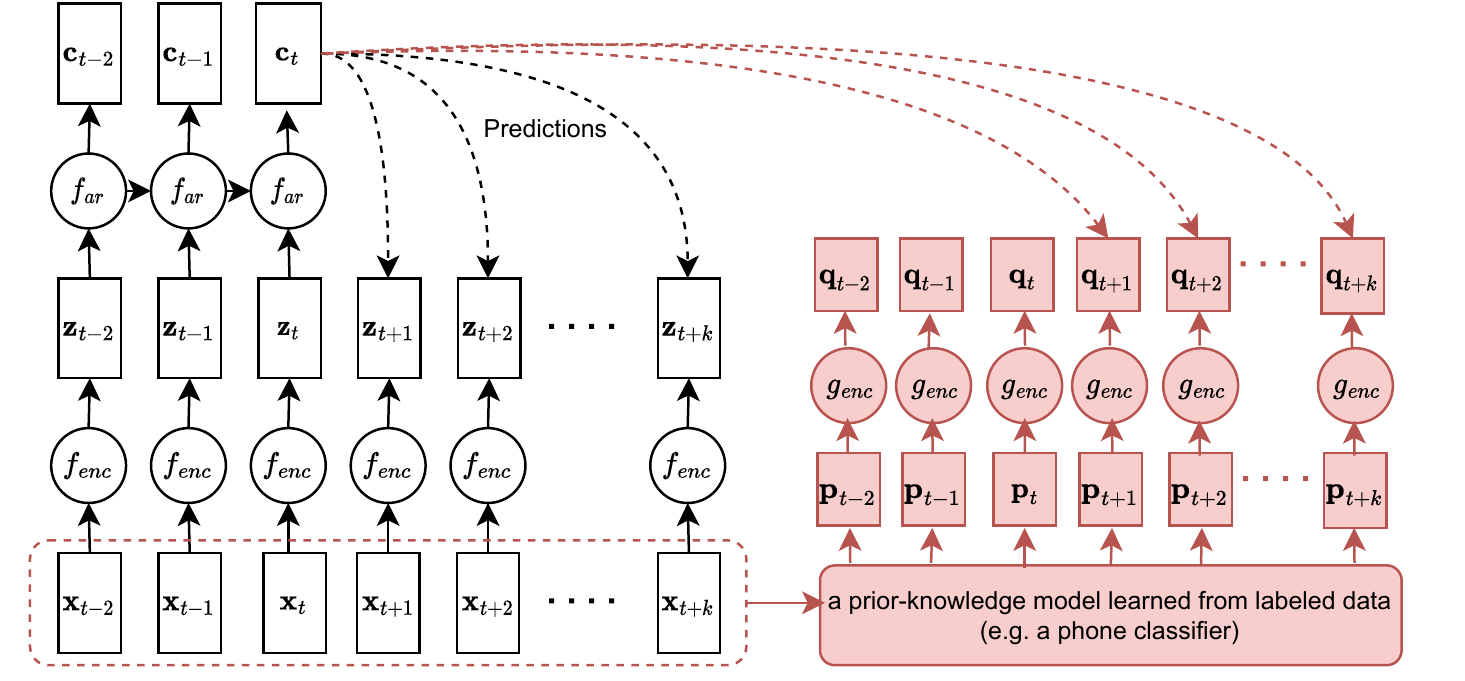}
  \caption{\small{\textit{Illustration of conventional Contrastive Predictive Coding (CPC) representation learning approach (left part) and our proposed Guided CPC (GCPC) method (right part in red). Parameters of the prior-knowledge model are fixed during training. $\mathbf{p}_{t}$ is a sequence of logits from a monophone classifier in our experiments.}}}
  \label{fig:GCPC}
\vspace{-3mm}
\end{figure}

The left part of Figure \ref{fig:GCPC} gives an overview of conventional CPC representation learning approach. Given frames of audio features $\mathbf{x}_t \in \mathcal{X}$, we first apply the feature encoder network $f_{enc}: \mathcal{X} \mapsto \mathcal{Z}$ to map the input sequence to a sequence of latent feature representations $\mathbf{z}_t \in \mathcal{Z} = f_{enc}(\mathbf{x}_t)$. An autoregressive context network $f_{ar}: \mathcal{Z} \mapsto \mathcal{C}$ summarizes all $\mathbf{z}_{\leq t}$ in the latent space and produces a contextual latent representation $\mathbf{c}_t = f_{ar}(\mathbf{z}_{\leq t})$

Both the feature encoder network and the autoregressive context network are trained to optimize the contrastive loss defined in Equation \ref{eq:regular_contrastive_loss_step} based on Noise-Contrastive Estimation (NCE) \cite{gutmann2010noise} for each step $k$, which equivalently maximizes the mutual information between $\mathbf{c}_t$ and the latent representation $\mathbf{z}_{t+k}$ that is $k$ steps in the future \cite{oord2018representation}.
\begin{equation}
\mathcal{L}_k=-\frac{1}{T-k}\sum_{t=1}^{T-k}\log\frac{\exp(\mathbf{z}_{t+k}^\top h_k(\mathbf{c}_t)/\kappa)} {\sum_{\mathbf{\tilde{z}\in \mathcal{Z}}}\exp(\mathbf{\tilde{z}}^\top h_k(\mathbf{c}_t)/\kappa)}
\label{eq:regular_contrastive_loss_step}
\end{equation}
where $\mathbf{\tilde{z}}$ is a set of negative samples sampled from the same audio example to represent the imposter distribution, $h_k(\mathbf{c}_t) = W_k\mathbf{c}_t+\mathbf{b}_k$ is a step-specific affine transformation applied to $\mathbf{c}_t$ for each step $k$, and $\kappa$ is the temperature. We optimize the final contrastive loss $\mathcal{L}_{C}$ by averaging $\mathcal{L}_k$ over the next K steps:
\begin{equation}
\mathcal{L}_C = \frac{1}{K}\sum_{k=1}^K \mathcal{L}_k
\label{eq:L_C_regular}
\end{equation}
\vspace{-6mm}


\subsection{Guided contrastive predictive coding model}
\label{sec:method:gcpc}
CPC learns representations from the complete data distribution, which may not be optimal for the downstream ASR task. In this paper, we propose to provide weak guidance for the contrastive loss. This weak guidance is provided in the form of posteriors from a prior-knowledge model learned from a small labeled dataset, and we use a monophone classifier for experimentation in the paper. As shown in the right part of Figure \ref{fig:GCPC}, we use an additional encoder network $g_{enc}: \mathcal{P} \mapsto \mathcal{Q}$ to map the sequence of unnormalized posteriors (logits) $\mathbf{p}_{t}$ to a sequence of latent representations $\mathbf{q}_{t} \in \mathcal{Q} = g_{enc}(\mathbf{p}_{t})$, and then optimize the guided contrastive loss $\mathcal{L}_C^{guided}$ defined in Equation \ref{eq:L_C_guided}.

\begin{equation}
\mathcal{L}_k^{guided}=-\frac{1}{T-k}\sum_{t=1}^{T-k}\log\frac{\exp(\mathbf{q}_{t+k}^\top h_k(\mathbf{c}_t)/\kappa)} {\sum_{\mathbf{\tilde{q}\in \mathcal{Q}}}\exp(\mathbf{\tilde{q}}^\top h_k(\mathbf{c}_t)/\kappa)}
\label{eq:guided_contrastive_loss_step}
\end{equation}

\begin{equation}
\mathcal{L}_C^{guided} = \frac{1}{K}\sum_{k=1}^K \mathcal{L}_k^{guided}
\label{eq:L_C_guided}
\end{equation}

During training, parameters of the prior-knowledge model are fixed. We hypothesize that representations $\mathbf{c}_{t}$ learned through this new technique could capture more phone discriminative characteristics since optimizing the guided contrastive loss helps maximizing the mutual information between $\mathbf{c}_{t}$ and transformation of phone posteriors $\mathbf{q}_{t+k}$. Thus, $\mathbf{c}_{t}$ might be more aligned with the downstream ASR task and serve as a better initialization point.
\vspace{-2mm}

\subsection{Contrastive pre-training for RNN-T ASR}
\label{sec:method:pretrain}
\begin{figure}[h!]
  \centering
  \begin{subfigure}[b]{0.56\linewidth}
    \includegraphics[width=\linewidth]{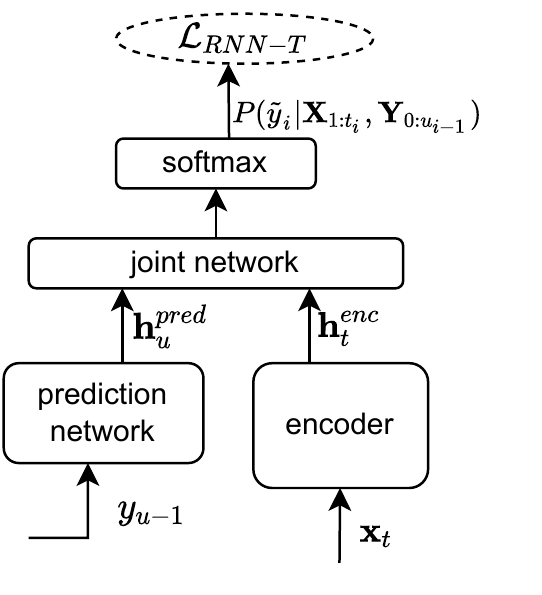}
    \caption{\small{RNN-T ASR}}
    \label{fig:RNN-T_general}
    \vspace{-2mm}
  \end{subfigure}
  \begin{subfigure}[b]{0.4\linewidth}
    \includegraphics[width=\linewidth]{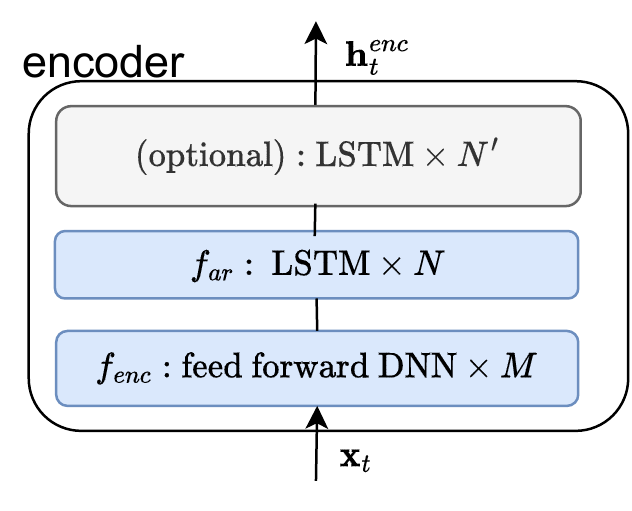}
    \caption{\small{encoder architecture}}
    \label{fig:RNN-T_encoder}
    \vspace{-2mm}
  \end{subfigure}
  \caption{\small{\textit{Overview of the RNN-T based ASR system and architecture of the encoder used for experimentation.}}}
  \label{fig:RNN-T}
  \vspace{-3mm}
\end{figure}

We use an RNN-T \cite{graves2012sequence} based ASR system for our experiments. The RNN-T model consists of an encoder, a prediction network and a joint network as shown in Figure \ref{fig:RNN-T_general}. Let $\mathcal{D}=\{(\mathbf{X}, \mathbf{Y})\}$ denote a single example from a training corpus where $\mathbf{X}=\{\mathbf{x}_1, \mathbf{x}_2,..., \mathbf{x}_T\}$ is a sequence of speech features and $\mathbf{Y}=\{y_1, y_2, ... y_U\}, y_u \in \mathcal{V}$ is a sequence of tokens from the vocabulary $\mathcal{V}$ (e.g. word pieces) representing the labels. The encoder maps each frame of the input speech features $\mathbf{x}_t$ to a hidden state $\mathbf{h}_t^{enc}$. The prediction network takes the embedding vector of the previous non-blank token $y_{u-1}$ and generates the hidden state $\mathbf{h}_u^{pred}$. The joint network is a feed-forward network that combines the outputs of the encoder and the prediction network to predict the conditional distribution over the next possible token $\tilde{y}_i \in \mathcal{V} \cup \langle blk\rangle$, where $ \langle blk\rangle$ denotes the blank symbol. The RNN-T loss is computed by marginalizing over all possible blank-augmented token sequences $\mathbf{\tilde{Y}}=\{\tilde{y}_1, \tilde{y}_2, ..., \tilde{y}_{T+U}\}$ aligned with each original token sequence $\mathbf{Y}$ and feature sequence $\mathbf{X}$:
\begin{equation}
\mathcal{L}_{RNN-T} = -\sum_{(\mathbf{X}, \mathbf{Y})\in \mathcal{D}}\log \sum_{\mathbf{\tilde{Y}}}
\prod_{i=1}^{T+U} P(\tilde{y}_i|\mathbf{X}_{1:t_i}, \mathbf{Y}_{0:u_{i-1}})
\label{eq:L_RNN-T}
\end{equation}
where the index $i$ in $\mathbf{\tilde{Y}}$ is mapped to the index $u_i$ in $\mathbf{Y}$ and the index $t_i$ in $\mathbf{X}$.

After pre-training with the CPC or GCPC approach, The feature encoder network $f_{enc}$ and the autoregressive context network $f_{ar}$ are used to initialize encoder of the RNN-T ASR model in our experiments as shown in Figure \ref{fig:RNN-T_encoder}. The remaining parts of the RNN-T model are randomly initialized before the final supervised training stage. When initializing the RNN-T encoder, we experiment with initializing the entire RNN-T encoder as well as initializing part of the RNN-T encoder. During the RNN-T supervised training stage, we additionally experiment with joint training combining the supervised RNN-T loss and the self-supervised contrastive loss.


\section{EXPERIMENTAL SETUP}
\label{sec:exp}
\subsection{Datasets}
We explore our approach on three datasets in different languages; two in-house far-field corpora of de-identified utterances in German and French from a voice assistant, and one public dataset, Librispeech \cite{panayotov2015librispeech} in English. When experimenting with the Librispeech data, we use the Libri-Light \cite{kahn2020libri} dataset for self-supervised pre-training. The numbers of hours for each dataset are summarized in Table \ref{table:dataset} below.

The test sets for the German and French ASR tasks consist of de-identified utterances from a voice assistant, similar to the training data. For the English ASR task, we report results on the Librispeech test-clean and test-other test sets. The amount of test data is also summarized in Table \ref{table:dataset}.

\begin{table}[h!]
\begin{center}
\caption{\small{\textit{Summary of datasets for experimentation.}}}
\vspace{-2mm}
\label{table:dataset}
\begin{tabular}{l|r|r | r }
\hline
\multirow{2}{*}{Language} & \multicolumn{2}{c}{Train (hrs)} & Evaluation\\
& unlabeled & labeled & (hrs)\\
\hline
German (in-house) & 142k & 25k & 15 \\
French (in-house) & 30k &  15k & 9.5 \\
English (Librispeech) & 60k & 960 & 5.4/5.1 \\
\hline
\end{tabular}
\end{center}
\vspace{-8mm}
\end{table}
\label{sec:exp:data}
\subsection{Model and training details}
\label{sec:exp:train}
The encoder of the RNN-T model for our experimentation contains 3 feed-forward (dense) layers of size 512 with ReLU non-linearity, followed by 8 LSTM layers with 1024 units for the German and French ASR tasks, and 6 LSTM layers with 1024 units for the Librispeech ASR task. For experiments that initialize the RNN-T encoder using CPC/GCPC pre-training methods, the feed-forward layers are initialized from the feature encoder network $f_{enc}$ of the CPC/GCPC pre-trained model illustrated in Figure \ref{fig:GCPC}, and the LSTM layers are initialized from the autoregressive context network $f_{ar}$. The RNN-T model also contains a single-layer LSTM prediction network with 1024 units, and a single-dense-layer joint network. When training the RNN-T ASR model, we use a sentence piece model \cite{kudo2018sentencepiece} containing 4000 sentence pieces which are trained with the corresponding monolingual dataset for each task.

The acoustic features used are 256-dimensional log short-time fourier transform (log-STFT), computed on a 25ms window with a frame shift of 10ms. The input log-STFT features from 3 consecutive frames are stacked for an effective frame size of 30ms, so that the final RNN-T input feature is of the dimension 768. There is no external language model used for any of the ASR tasks.

When computing the guided contrastive loss, extra dense layers $g_{enc}$ are added on top of phone logits, and are updated during the pre-training stage. Our ablation studies described in Section \ref{sec:ablation:hyperparams} will demonstrate the impact of these extra feed-forward layers. The number of steps $K$ (in Equation \ref{eq:L_C_regular}) used for pre-training is 4. 
\vspace{-4mm}
\section{Results}
\subsection{Phone Classification}
\label{sec:results:phone_classifiers}
The phone classification model is a 5-layer LSTM model with 768 units in each layer. It is trained with the standard cross-entropy loss. Frame-level monophone targets for the internal datasets are generated from the 1-best decoding output of a hybrid LSTM-HMM ASR model. For Librispeech, we obtain the frame-level phone targets using the Montreal forced aligner\footnote{https://github.com/CorentinJ/librispeech-alignments}. Frame-level accuracy for these phone classifiers are shown in Table \ref{table:phone_classifier}. These phone classifiers are then used as the prior knowledge models for the guided contrastive loss based pre-training. 

\begin{table}[h!]
\begin{center}
\caption{\small{\textit{Frame-level phone accuracy (monophones) for the phone classifier built for each language.}}}
\label{table:phone_classifier}
\vspace{-2mm}
\begin{tabular}{c|c| c}
\hline
Language & \# of phones & Accuracy (\%)\\
\hline
German                    &   55 & 75.94 \\
French                    &   45 & 81.39\\
Librispeech dev-clean     &   72 & 80.58 \\
\hline
\end{tabular}
\end{center}
\vspace{-3mm}
\end{table}
\vspace{-4mm}

\subsection{ASR results}
\label{sec:results:asr_tasks}
The results for each ASR task (German, French and English) with different encoder pre-training methods are reported in Tables \ref{table:ASR_WERR} and \ref{table:ASR_WER_libri}. The baseline RNN-T ASR models ($B_{G_1}$, $B_{F_1}$, $B_{L_1}$) are trained from scratch. Encoders of $M_{G_2}$, $M_{F_2}$ and $M_{L_2}$ are initialized with the encoder pre-trained with standard CPC method explained in Section \ref{sec:method:cpc}. Encoders of $M_{G_3}$, $M_{F_3}$ and $M_{L_3}$ are initialized with our proposed Guided CPC (GCPC) method explained in Section \ref{sec:method:gcpc}. For a more comprehensive comparison, we also pre-train the encoder with the cross-entropy loss using phone posteriors (PCE) obtained from the phone classifier described in Section \ref{sec:results:phone_classifiers}, and these results are reported for models $M_{G_1}$, $M_{F_1}$, $M_{L_1}$. Note we only report relative WER reduction (WERR) on tasks using the internal data shown in Table \ref{table:ASR_WERR}, but we do report the absolute WER on the Librispeech task shown in Table \ref{table:ASR_WER_libri}.
\begin{table}[h!]
\begin{center}
\caption{\small{\textit{Relative Word Error Rate Reduction (WERR) w.r.t RNN-T ASR baseline when using different encoder pre-training methods on German and French ASR tasks. Negative WERR indicates a degradation. Best numbers in bold.}}}
\vspace{-2mm}
\label{table:ASR_WERR}
\begin{tabular}{c|c| rr}
\hline
\multirow{2}{*}{Model} & RNN-T encoder        & \multirow{2}{*}{WERR\%} \\
                          &initialization &  &  \\
\hline
\multicolumn{2}{l}{\textit{German ASR task}} & Test German & \\
\hline
$B_{G_1}$\tablefootnote{Table Notation: B is baseline, M is experimental model, letter in subscript (G/F/E) is language, and numeral in the subscript is the experiment id. The extra letter in subscripts in Section \ref{sec:results:ablation_studies} refers to the ablation study id.} & -                            & 0.00 & \\
$M_{G_1}$ & PCE                          & 2.11 & \\
$M_{G_2}$ & CPC                          & 2.96 & \\
$M_{G_3}$ & GCPC                         & \textbf{4.44} & \\
\hline
\multicolumn{2}{l}{\textit{French ASR task}} & Test French &\\
\hline
$B_{F_1}$ & -                            & 0.00 &  \\
$M_{F_1}$ & PCE                          & -0.50 &  \\
$M_{F_2}$ & CPC                          & 1.01 &  \\
$M_{F_3}$ & GCPC                         & \textbf{6.55} &  \\
\hline
\end{tabular}
\end{center}
\vspace{-5mm}
\end{table}

Overall, the standard contrastive pre-training on the RNN-T encoder reduces WER on our internal German and French ASR tasks by 2.96\% and 1.01\% respectively relative to the baseline, while our proposed guided contrastive pre-training method brings higher relative WER reductions (4.44\% and 6.55\% on the German and French ASR tasks respectively). The standard phone cross-entropy pre-training leads to a worse WER than both the CPC and GCPC pre-training, which indicates the importance of the contrastive term in the pre-training loss function. On the Librispeech task, we observe similar trend under the Test-Clean condition. However, under the Test-Other condition, we don't see benefits of the GCPC pre-training method. From all our experimental results, we observe that for the tasks with larger labeled training dataset, the gain with the pre-training techniques are smaller. 

\begin{table}[h!]
\begin{center}
\caption{\small{\textit{Word Error Rate (WER) when using different encoder pre-training methods on Librispeech ASR task. Best numbers in bold.}}}
\vspace{-2mm}
\label{table:ASR_WER_libri}
\begin{tabular}{c|c| l@{\hskip -3mm}l}
\hline
\multirow{2}{*}{Model} & RNN-T encoder        & \multirow{2}{*}{WER\% (WERR\%)}\\
                          &initialization &  &  \\
\hline
\multicolumn{2}{l}{\textit{Librispeech ASR task}} & Test-Clean & Test-Other\\
\hline
$B_{L_1}$ & -                           & 6.74 & 17.44  \\
$M_{L_1}$ & PCE                         & 7.22 (-7.12) & 19.25 (-10.38)  \\
$M_{L_2}$ & CPC                         & 5.77 (14.39)  & \textbf{16.05 (7.97)}  \\
$M_{L_3}$ & GCPC                        & \textbf{5.70 (15.43)} &  16.21 (7.05) \\
\hline
\end{tabular}
\end{center}
\vspace{-6mm}
\end{table}
\vspace{-3mm}
\subsection{Ablation studies}
\label{sec:results:ablation_studies}
In order to identify the best training scheme, we perform ablation studies on the internal German dataset. We only show results for the German ASR task, but the WER trends for the ASR models for all 3 languages on both the development and test data were similar. The training scheme was tuned on a held out development set, but we show our results on the test set so the numbers in sections \ref{sec:results:asr_tasks} and \ref{sec:results:ablation_studies} are comparable.
\vspace{-2mm}
\subsubsection{Two-stage training versus joint training}
Since more recent research demonstrates that joint training with both supervised and self-supervised losses can directly optimize the ASR performance \cite{bai2022joint}, we experiment with joint training combining the supervised RNN-T loss $\mathcal{L}_{RNN-T}$ and the self-supervised contrastive loss $\mathcal{L}_C$ as shown in Table \ref{table:joint}. Note that we use a different baseline $B_{G2}$ with a relatively small batch size due to memory limitations of the GPU devices used for experimentation. We demonstrate that joint training from scratch ($M_{G_{A1}}$) brings a relative WERR of 3.32\%, which is slightly better than the 2.96\% relative WERR obtained from the conventional two-stage training scheme ($M_{G_2}$ in Table \ref{table:ASR_WERR}). On top of this, contrastively pre-training the RNN-T encoder ($M_{G_{A2}}$) doesn't seem to further improve the WER. Considering similar WERRs from these two training schemes and the high memory consumption by the joint training which causes training instability, we use the conventional two-stage training scheme for our proposed method.

\begin{table}[h!]
\begin{center}
\caption{\small{\textit{Relative Word Error Rate Reduction (WERR) w.r.t RNN-T ASR baseline with joint training on the German ASR task.}}}
\vspace{-2mm}
\label{table:joint}
\begin{tabular}{c|c| c|rr}
\hline
\multirow{2}{*}{Model} & RNN-T encoder        &  \multirow{2}{*}{Loss}& WERR\% \\
                          &initialization &      & Test German \\
\hline
$B_{G_2}$ & random   & \small{$\mathcal{L}_{RNN-T}$}                           & 0.00  \\
$M_{G_{A1}}$ & random & \small{$\mathcal{L}_{RNN-T} + \mathcal{L}_C$}          & 3.32  \\
$M_{G_{A2}}$ & CPC & \small{$\mathcal{L}_{RNN-T} + \mathcal{L}_C$}             & 1.95  \\
\hline
\end{tabular}
\end{center}
\end{table}
\vspace{-5mm}

\begin{table*}[t]
\begin{center}
\caption{\small{\textit{Relative Word Error Rate Reduction (WERR) shown w.r.t RNN-T ASR baseline when experimenting with different pre-training stages on the German ASR task.}}}
\vspace{-2mm}
\label{table:pretrain_schemes}
\begin{tabular}{l|ccc|ccc|rr}
\hline
\multirow{3}{*}{Model} & \multicolumn{6}{c|}{RNN-T encoder} &          \multicolumn{2}{c}{WERR\%} \\
\cline{2-9}
                          &  \multicolumn{3}{c|}{($f_{enc} + f_{ar}$)} & \multicolumn{3}{c|}{remaining layers} & \multirow{2}{*}{Test German}&\\
\cline{2-7}
                          & architecture & initialization & trainable  & architecture & initialization &  trainable &  &  \\ 
\hline
$B_{G_1}$        &$\text{DNN}\times3+\text{LSTM}\times8$     & random      & Yes & - & - &-       & 0.00 & \\
$M_{G_1}$        &$\text{DNN}\times3+\text{LSTM}\times8$     & CPC      & Yes & - & - &-       &  \textbf{2.96} &  \\
$M_{G_{B1}}$        &$\text{DNN}\times3+\text{LSTM}\times2$     & CPC      & No  & $\text{LSTM}\times6$ & random & Yes       &-10.36  &  \\
$M_{G_{B2}}$        &$\text{DNN}\times3+\text{LSTM}\times2$     & CPC       & yes  & $\text{LSTM}\times6$ & random  & Yes       &-0.85  &  \\
$M_{G_{B3}}$        &$\text{DNN}\times3+\text{LSTM}\times2$     & $M_{G_{B1}}$       & yes  & $\text{LSTM}\times6$ & $M_{G_{B1}}$  & Yes       &  -1.69&  \\
\hline
\end{tabular}
\vspace{-5mm}
\end{center}
\end{table*}

\subsubsection{Optimizing the pre-training stage}
All the pre-training experiments reported in Section \ref{sec:results:asr_tasks} utilize an RNN-T encoder with all its layers pre-trained. We also experiment with different pre-training stages and the resulting WERRs are shown in Table \ref{table:pretrain_schemes}. For model $M_{G_{B1}}$, we only pre-train the RNN-T encoder up to the second LSTM layer with the contrastive loss. We then freeze those pre-trained layers and randomly initialize the rest 6 LSTM layers during the supervised RNN-T training stage. The supervised training stage for model $M_{G_{B2}}$ is similar to that of model $M_{G_{B1}}$, except that we don't freeze the pre-trained layers. Model $M_{G_{B3}}$  further tunes model $M_{G_{B1}}$ by making all layers trainable. According to WERRs reported in Table \ref{table:pretrain_schemes}, pre-training the entire RNN-T encoder gives the best WER. Therefore, we adopt this method of pre-training for all other experiments in the paper. 

\subsubsection{Hyperparameter tuning for contrastive pre-training}
\label{sec:ablation:hyperparams}

We tune two hyperparameters for guided contrastive pre-training. The first hyperparameter is the temperature for the loss function, $\kappa$. Second, instead of using phone logits directly to compute the contrastive loss, we experiment with adding trainable feed-forward layers ($g_{enc}$ illustrated in Figure \ref{fig:GCPC}) on top of the phone logits before computing the contrastive loss. The intuition behind this is to let the model learn derivative features from the phone posteriors, which might be more suitable for the downstream ASR task. 
Relative WERRs with different hyperparameters are shown in Table \ref{table:hyperparams}. We note that when we use phone logits directly as latent representations, the WER degrades by 16.49\% relative to the baseline ($M_{G_{C7}}$ vs $B_{G_1}$). Using the learnable feed-forward layers $g_{enc}$ to generate latent representations is critical to the GCPC pre-training. Based on the tuning results, using 2 learnable feed-forward layers to generate  latent representations for contrastive learning gives the largest WER reduction. Finally, the tuning of the temperature parameter $\kappa$ shows that 0.01 is the optimal value for the German ASR GCPC pre-training, and 0.1 is the optimal value for the conventional CPC pre-training. Note that we experimented with the outputs of the intermediate layer of the phone classifier as inputs to compute contrastive loss and found that they lead to a worse ASR performance compared to logits followed by trainable feed-forward layers (results not in the Table).

\begin{table}[h!]
\begin{center}
\caption{\small{\textit{Relative Word Error Rate Reduction (WERR) shown w.r.t RNN-T ASR baseline when tuning hyperparameters for contrastive pre-training on the German ASR task.}}}
\vspace{-2mm}
\label{table:hyperparams}
\begin{tabular}{c|c|c|c| r}
\hline
\multirow{2}{*}{Model} & \multicolumn{3}{c|}{RNN-T encoder initialization}     & \multirow{2}{*}{WERR\%} \\
                          &method & $\kappa$  & $g_{enc}$ &  \\
\hline
$B_{G_1}$ & -   & - &  -                        & 0.00  \\
$M_{G_1}$ & CPC   & 0.1 &  -                        & 2.96  \\
$M_{G_{C1}}$ & CPC   & 0.02 &  -                        & 0.42  \\
$M_{G_{C2}}$ & CPC   & 0.01 &  -                        & -1.06  \\
$M_{G_{C3}}$ & GCPC & 0.1 &  $\text{DNN}\times2$        & -18.18   \\
$M_{G_{C4}}$ & GCPC& 0.02 & $\text{DNN}\times2$                           & 3.59  \\
$M_{G_{3}}$ & GCPC & 0.01 &  $\text{DNN}\times2$                         &  \textbf{4.44}   \\
$M_{G_{C5}}$ & GCPC & 1e-5 &  $\text{DNN}\times2$                         & 1.06   \\
$M_{G_{C6}}$ & GCPC& 0.02 & $\text{DNN}\times3$                           & 2.54   \\
$M_{G_{C7}}$ & GCPC & 0.02 &  None                         & -16.49   \\
\hline

\end{tabular}
\end{center}
\vspace{-5mm}
\end{table}
\subsubsection{Joint contrastive pre-training}
We perform an additional experiment where the pre-training objective $\mathcal{L}_C^{joint}$ consists of both the regular contrastive loss ($\mathcal{L}_{C}$ defined in Equation \ref{eq:L_C_regular}) and the guided contrastive loss ($\mathcal{L}_{C}^{guided}$ defined in Equation \ref{eq:L_C_guided}). 
\begin{equation}
    \mathcal{L}_C^{joint} = \mathcal{L}_{C} + \mathcal{L}_{C}^{guided}
\end{equation}
According to results in Table \ref{table:joint_pretrain}, we can see that guided contrastive pre-training performs better than both regular contrastive as well as the joint contrastive pre-training in terms of the final WER.

\begin{table}[h!]
\begin{center}
\caption{\small{\textit{Relative Word Error Rate Reduction (WERR) w.r.t RNN-T ASR baseline when using individual contrastive loss and joint contrastive loss for pre-training on the German ASR task.}}}
\vspace{-2mm}
\label{table:joint_pretrain}
\begin{tabular}{c|c| r}
\hline
\multirow{2}{*}{Model} & RNN-T encoder        & WERR\% \\
                          &initialization &  \\
\hline
$B_{G_1}$ & -                            & 0.00  \\
$M_{G_{2}}$ & CPC                          & 2.96  \\
$M_{G_{3}}$ & GCPC                         & \textbf{4.44} \\
$M_{G_{D1}}$ & CPC+GCPC                     & 1.69  \\
\hline
\end{tabular}
\end{center}
\vspace{-8mm}
\end{table}

\subsection{Analysis of representations}

\begin{figure*}[h]
\begin{subfigure}{0.66\textwidth}
\begin{subfigure}{0.5\textwidth}
\centering
  \includegraphics[width=0.87\linewidth]{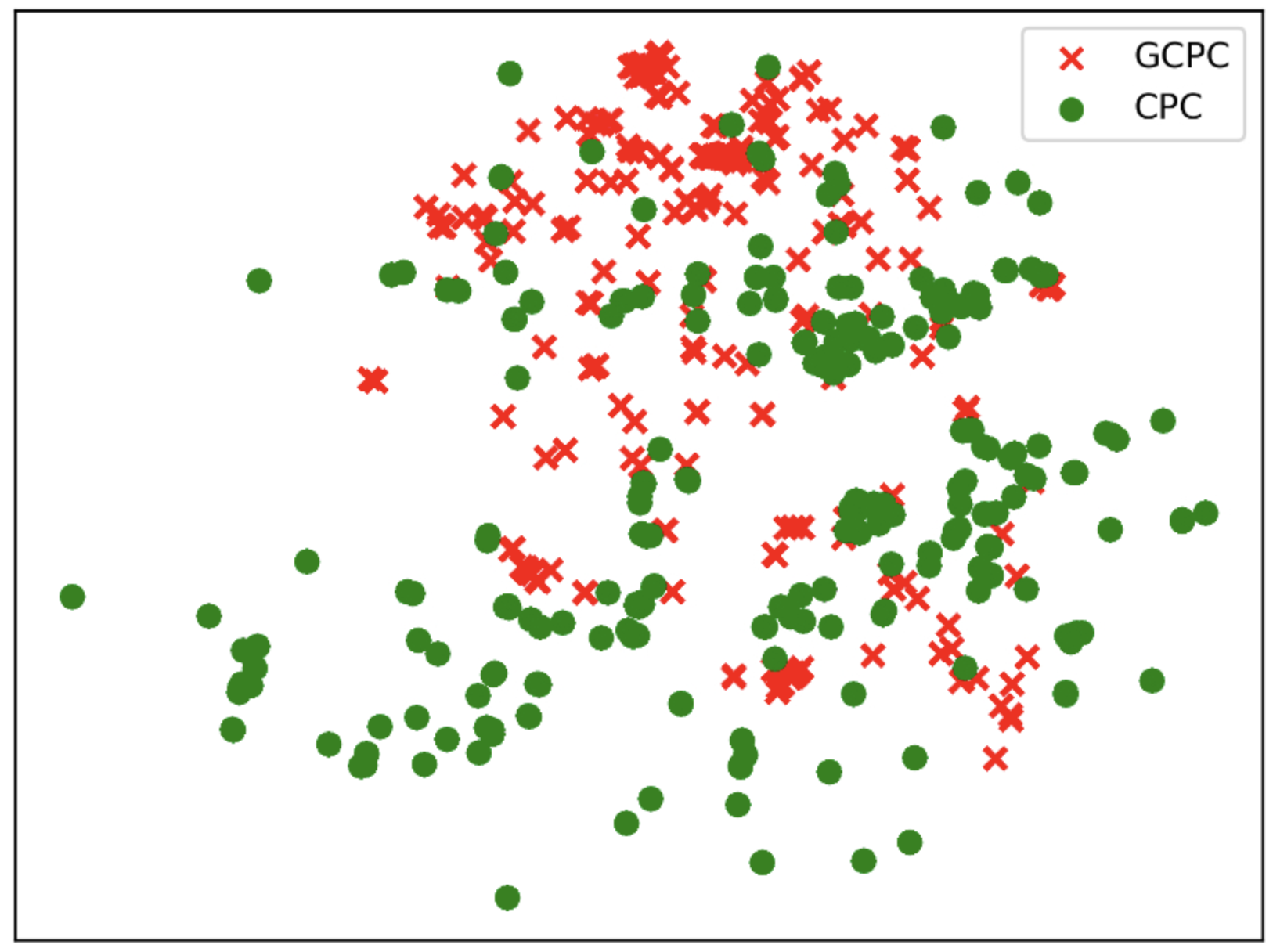}\quad
  \caption{Phone a: - Unrounded vowel}
  \label{fig:sfig1}
\end{subfigure}
\begin{subfigure}{0.5\textwidth}
\centering
  \includegraphics[width=0.87\linewidth]{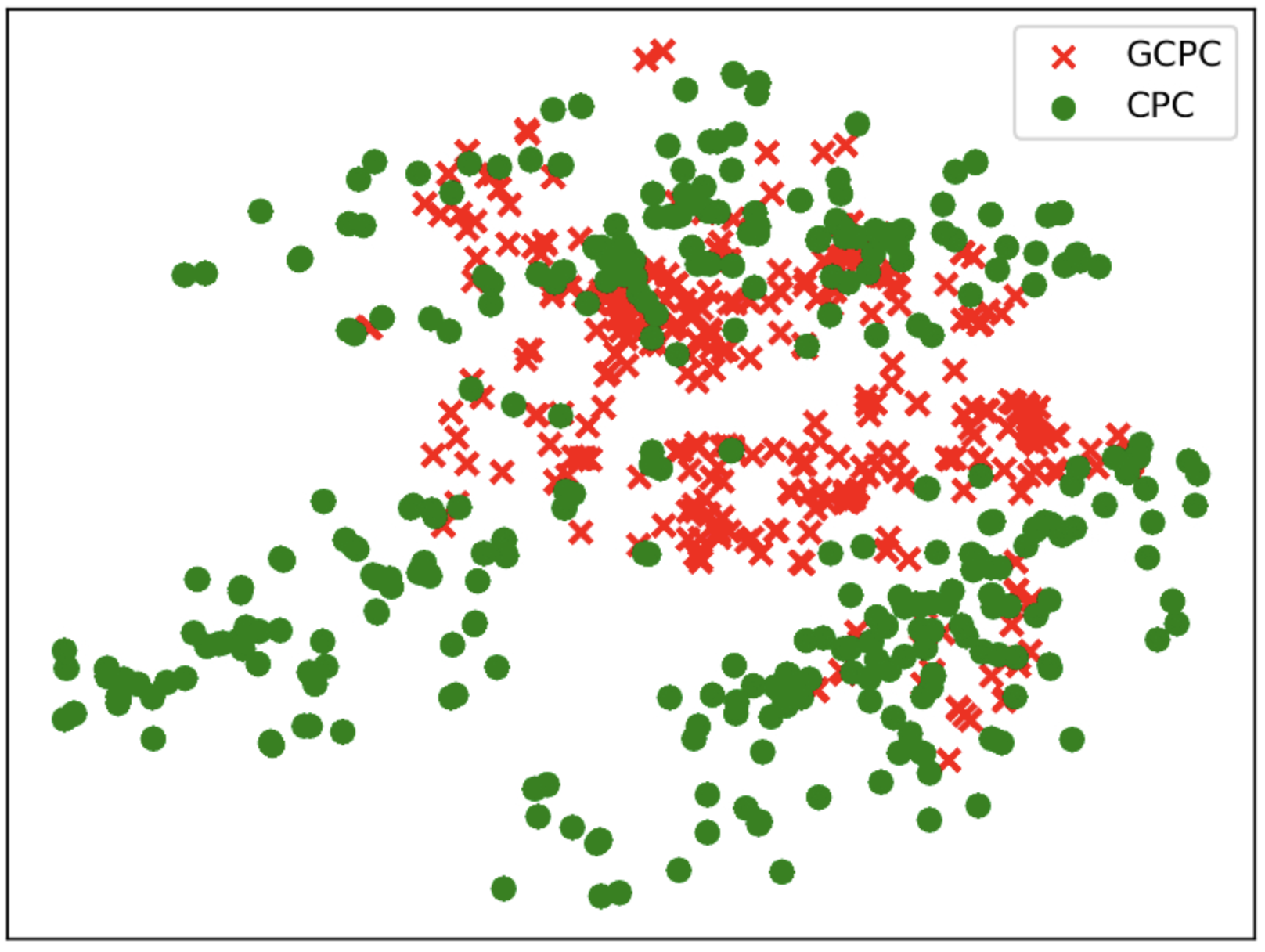}
  \caption{Phone I- Unrounded vowel}
  \label{fig:sfig2}
\end{subfigure}%
\hfill
\begin{subfigure}{0.5\textwidth}
\centering
  \includegraphics[width=0.87\linewidth]{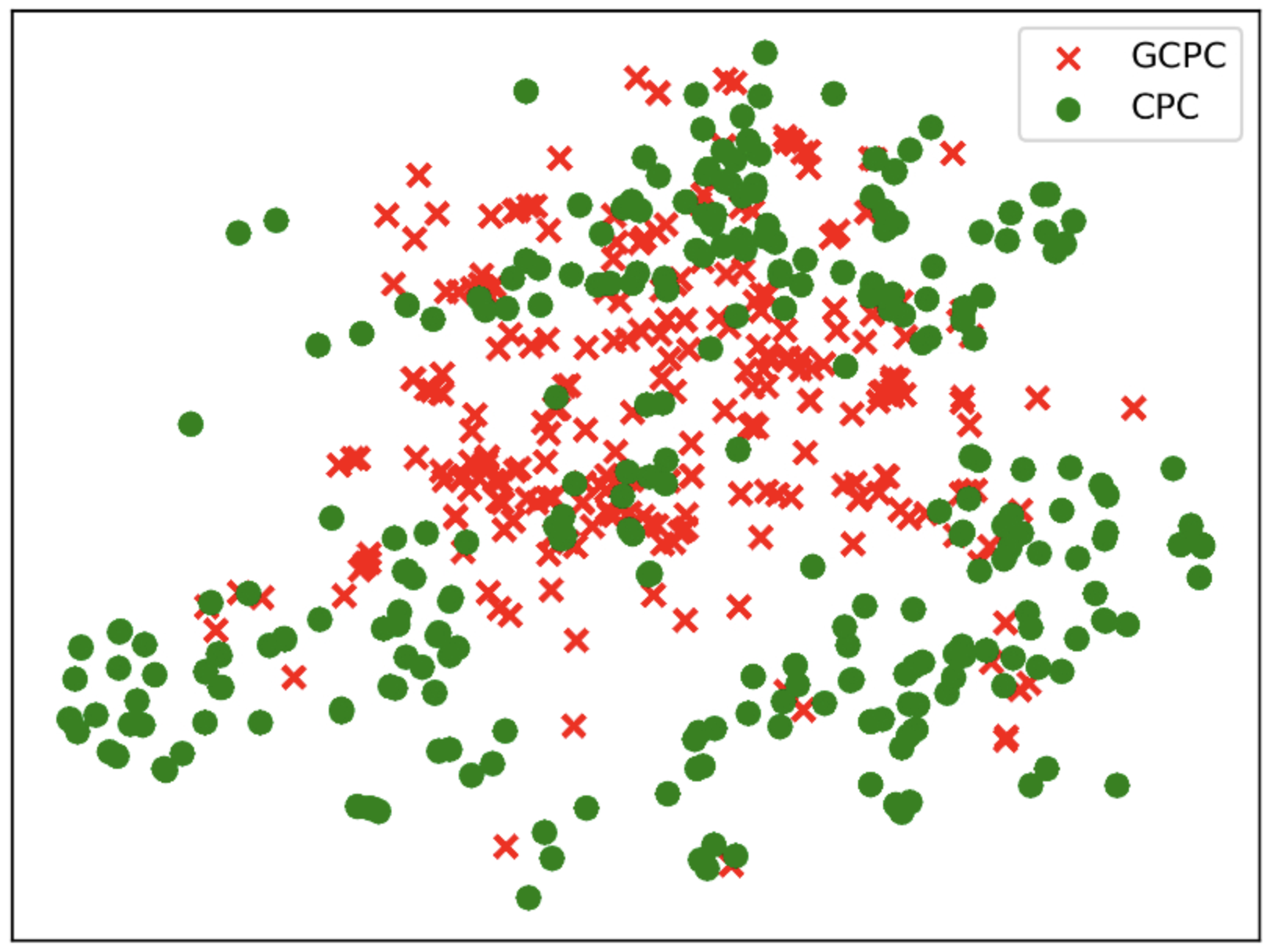}\quad
  \caption{Phone d-voiced alveolar plosive}
  \label{fig:sfig3}
\end{subfigure}
\begin{subfigure}{0.5\textwidth}
\centering
  \includegraphics[width=0.87\linewidth]{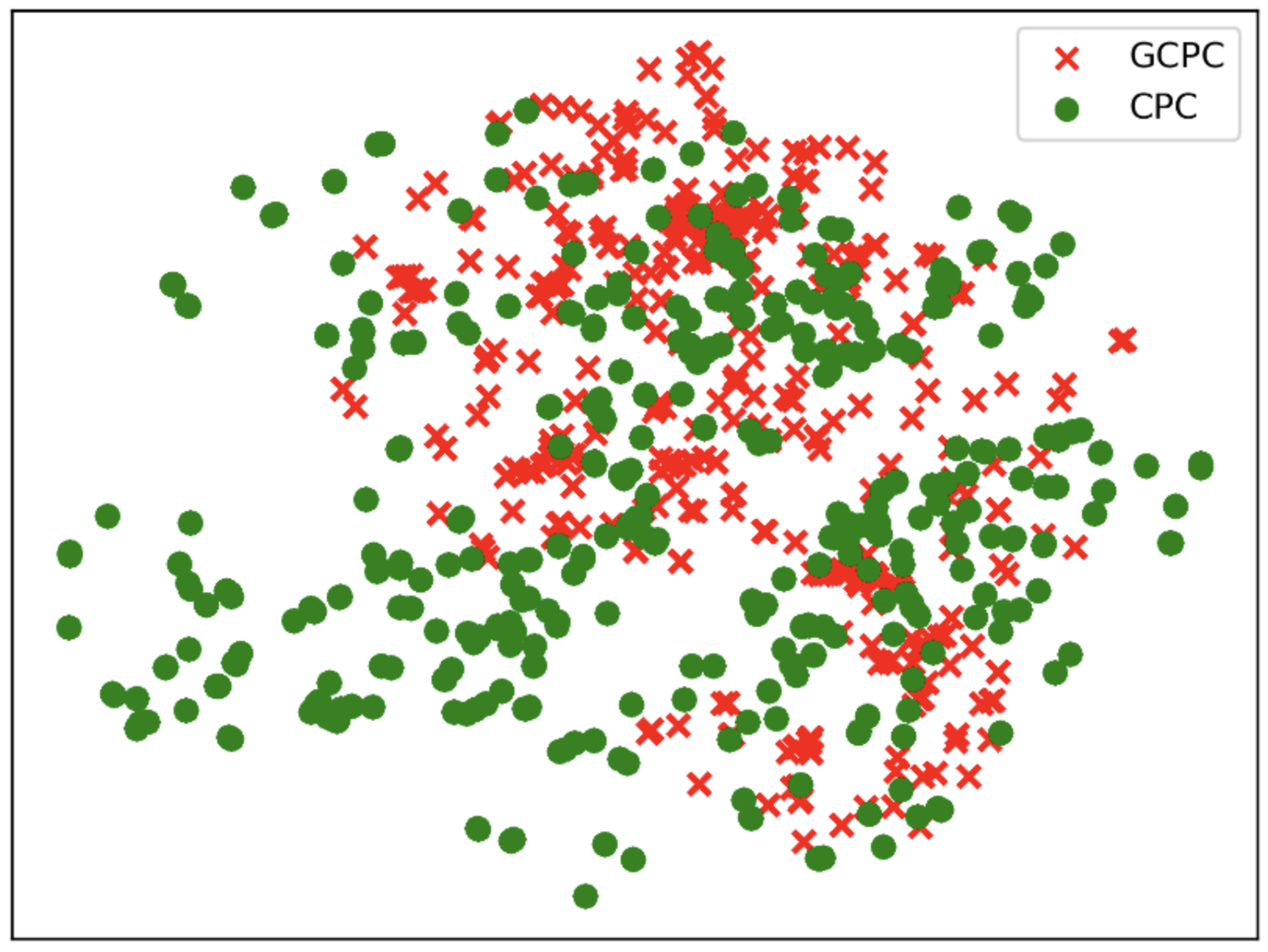}
  \caption{Phone N-Velar nasal}
  \label{fig:sfig4}
\end{subfigure}
\end{subfigure}%
\rulesep
\begin{subfigure}{0.33\textwidth}
\begin{subfigure}{\textwidth}
\centering
  \includegraphics[width=0.87\linewidth]{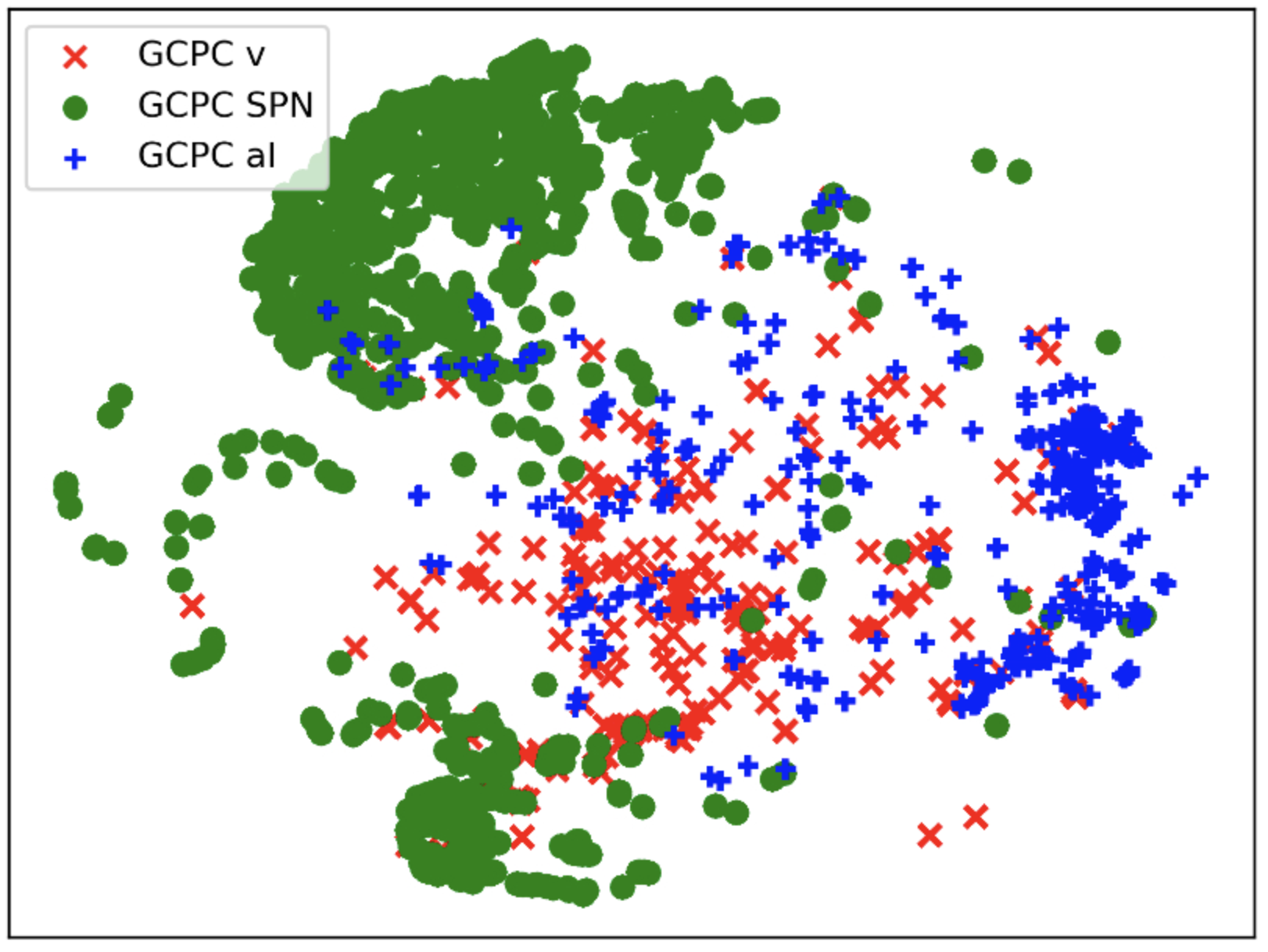}
  \caption{GCPC}
  \label{fig:sfig5}
\end{subfigure}%
\hfill
\begin{subfigure}{\textwidth}
\centering
  \includegraphics[width=0.87\linewidth]{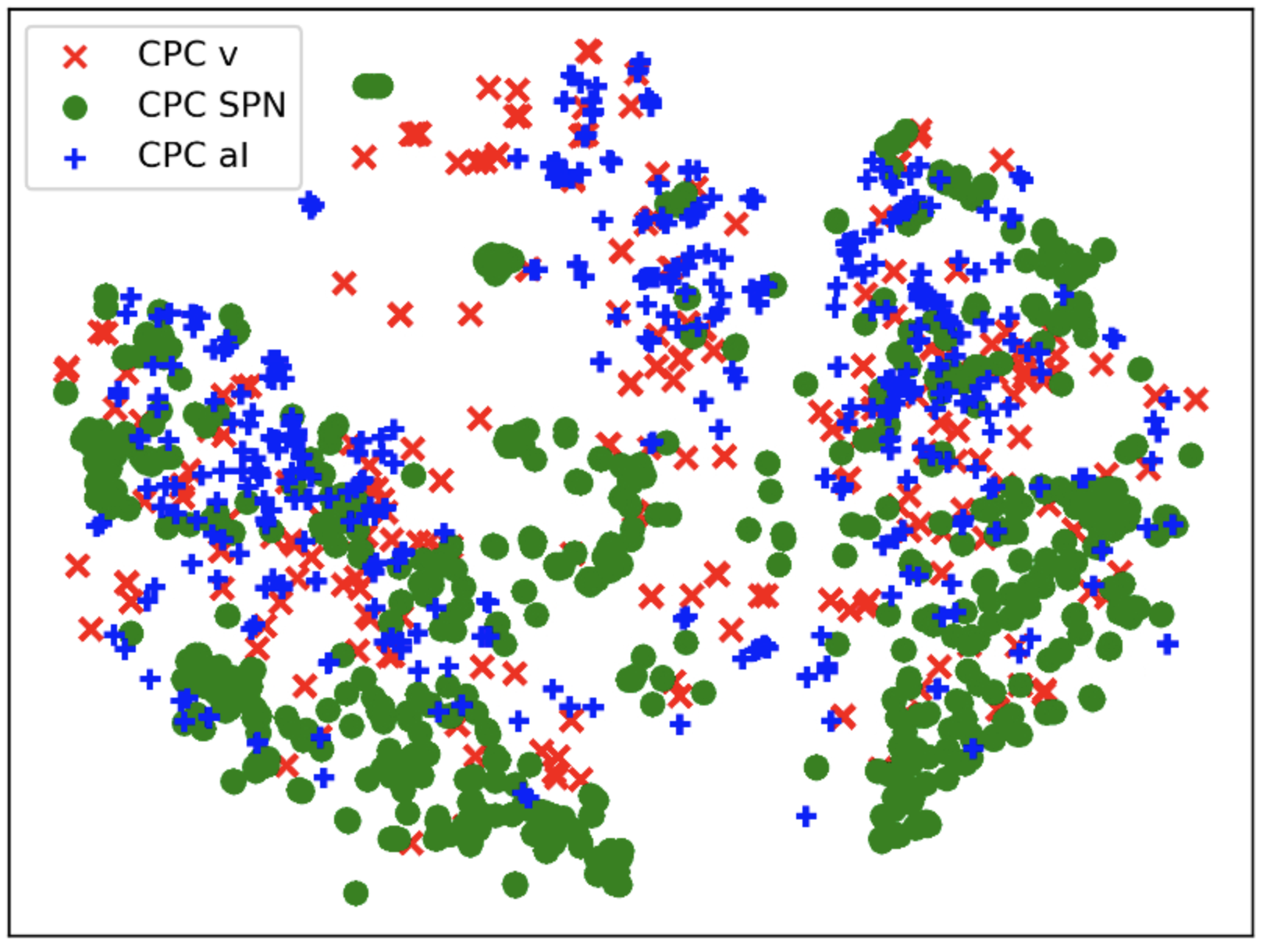}
  \caption{CPC}
  \label{fig:sfig6}
\end{subfigure}%
\end{subfigure}%
\vspace{-2mm}
\caption{\ref{fig:sfig1}-\ref{fig:sfig4} show  the t-SNE visualizations of 4 German phones from CPC and guided CPC pre-trained RNN-T encoders. \ref{fig:sfig5}-\ref{fig:sfig6} present the t-SNE visualizations for 3 phones on the same plot (v, aI and SPN (speech like noise)) obtained from GCPC and CPC pre-trained RNN-T encoders respectively, indicating better phone separation with GCPC pre-trained encoder.}
\label{fig:TSNE}
\vspace{-4mm}
\end{figure*}

To better understand the gains from using the guided contrastive loss, we visualize the output of the RNN-T encoder pre-trained using various methods with the German data. We take the frame level representations for a subset of the development set from the pre-trained RNN-T encoder and apply the t-Distributed Stochastic Neighbor Embedding (t-SNE) algorithm \cite{van2008visualizing} to reduce the embedding dimension from
1024 to 2. We plot these 2-dimensional embeddings obtained from CPC and guided CPC per monophone. 

Figure \ref{fig:sfig1}-\ref{fig:sfig4} show the t-SNE visualizations for four different monophones for the German ASR task. We use the X-SAMPA phone-set for the experiments. For the visualization, we choose two vowels, one plosive and one nasal phone. From these figures, we observe that the representations for each of the phone classes are more closely clustered when using the guided CPC pre-training as compared to using traditional CPC pre-training. Figure \ref{fig:sfig5}-\ref{fig:sfig6} show the t-SNE visualization for three different phones on the same plot from CPC and GCPC pre-trained encoders. Although there is overlap between the different phone clusters, we observe better separation between frames from the different phones with the GCPC pre-trained encoder as compared to the CPC pre-trained encoder. The separation is more pronounced between voiced frames compared to frames with speech like noise (SPN). Table \ref{table:breakdown} breaks down the relative reduction by substitution, deletion and insertion errors on the German ASR task. The breakdown shows that CPC pre-training degrades the insertion error rate by 3.45\% relative, while GCPC pre-training improves the insertion error rate by 3.39\%. The better separation between frames with speech like noise and other speech frames explains this improvement in insertion errors.

\begin{table}[h!]
\begin{center}
\caption{\small{\textit{Relative Error Rate Reduction in substitution (SUBR), insertion (INSR) and deletion (DELR) errors w.r.t RNN-T ASR baseline when using different pre-training methods for the German ASR Task.}}}
\label{table:breakdown}
\vspace{-2mm}
\begin{tabular}{c|c|r|r|r}
\hline
\multirow{2}{*}{Model} & RNN-T encoder        & SUBR\% & INSR\% & DELR\% \\
                          &initialization &  & & \\
\hline
$B_{G_1}$ & -                            & 0.00 & 0.00 & 0.00\\
$M_{G_2}$ & CPC                          & 1.45 & -3.45 & 8.57 \\
$M_{G_3}$ & GCPC                         & 2.55 & 3.39 & 9.29 \\
\hline
\end{tabular}
\end{center}
\vspace{-5mm}
\end{table}

\section{Conclusion}
We have shown that injecting prior knowledge in the form of phone posteriors during the contrastive pre-training stage helps improve the performance of three downstream ASR tasks compared to applying regular contrastive pre-training. On the German and French ASR tasks, our method gives a 4.44\% and 6.55\% relative WERR respectively, while the  regular CPC method just brings a 2.96\% and 1.01\% relative WERR. On the Librispeech ASR task (test-clean), our pre-training method reduces the WER relatively by 15.43\% compared to training the ASR model from scratch. From the t-SNE visualizations of the pre-trained embeddings using CPC and guided CPC methods, we observe closer clustering among frames belonging to the same phone with guided contrastive pre-training, and better separation between speech frames and frames with speech-like noise, leading to higher reduction in insertion error rates with the guided contrastive pre-training.

\clearpage
\newpage
\bibliographystyle{IEEEbib}
\bibliography{strings,refs}

\end{document}